\documentclass[letterpaper, 10 pt, conference]{ieeeconf}
\IEEEoverridecommandlockouts    
\usepackage{amsmath} 
\usepackage{amssymb}  
\usepackage{float}
\usepackage{graphicx}
\usepackage{booktabs}
\usepackage{subcaption}
\usepackage{caption}
\usepackage{tabularx}
\usepackage{multirow}
\usepackage{collcell}
\usepackage{color, xcolor, colortbl}
\definecolor{Gray}{gray}{0.9}
\definecolor{DarkGray}{gray}{0.8}
\definecolor{DarkerGray}{gray}{0.7}
\usepackage{balance}
\usepackage{tikz}
\usepackage{hyperref}

\newcommand*{\opacity}{40}
\definecolor{high}{HTML}{32D732}
\definecolor{mid}{HTML}{FFBF00}
\definecolor{low}{HTML}{FF0000} 
\newcommand*{\minval}{0.00}
\newcommand*{\midval}{0.50}
\newcommand*{\maxval}{1.00}
\newcommand{\gradient}[1]{
    \ifdim #1 pt > \midval pt
            \pgfmathparse{int(round(100*(#1/(\maxval-\midval))-(\midval*(100/(\maxval-\midval)))))}
            \xdef\tempa{\pgfmathresult}
            \cellcolor{high!\tempa!mid!\opacity} #1
        \else
            \pgfmathparse{int(round(100*(#1/(\midval-\minval))-(\minval*(100/(\midval-\minval)))))}
            \xdef\tempa{\pgfmathresult}
            \cellcolor{mid!\tempa!low!\opacity} #1
        \fi
        }

\title{\LARGE \bf
Intrinsic Language-Guided Exploration for \\ Complex Long-Horizon Robotic Manipulation Tasks
}

\author{Eleftherios Triantafyllidis$^{1}$, Filippos Christianos$^{1}$ and Zhibin Li$^{1,2}$
\thanks{$^{1}$Authors are with the School of Informatics, The University of Edinburgh, United Kingdom. {\tt\footnotesize \{eleftherios.triantafyllidis, f.christianos\}@ed.ac.uk}}%
\thanks{$^{2}$Zhibin Li is with the Department of Computer Science, University College London, United Kingdom. {\tt\footnotesize alex.li@ucl.ac.uk}}%
}

\begin{document}
\maketitle
\thispagestyle{empty}
\pagestyle{empty}

\begin{abstract}
Current reinforcement learning algorithms struggle in sparse and complex environments, most notably in long-horizon manipulation tasks entailing a plethora of different sequences. In this work, we propose the Intrinsically Guided Exploration from Large Language Models (IGE-LLMs) framework. By leveraging LLMs as an assistive intrinsic reward, IGE-LLMs guides the exploratory process in reinforcement learning to address intricate long-horizon with sparse rewards robotic manipulation tasks. We evaluate our framework and related intrinsic learning methods in an environment challenged with exploration, and a complex robotic manipulation task challenged by both exploration and long-horizons. Results show IGE-LLMs (i) exhibit notably higher performance over related intrinsic methods and the direct use of LLMs in decision-making, (ii) can be combined and complement existing learning methods highlighting its modularity, (iii) are fairly insensitive to different intrinsic scaling parameters, and (iv) maintain robustness against increased levels of uncertainty and horizons. 
\end{abstract}

\section{INTRODUCTION}
Current deep Reinforcement Learning (RL) approaches are faced with significant challenges in complex and sparse environments, particularly with long-horizon sequential tasks in robotic manipulation \cite{davchev2021wish, 8843293, roman}. An essential challenge in RL is the need for exploration, where immediate feedback from the environment is not readily apparent and usually in the form of sparse rewards \cite{10.5555/3535850.3535978}. Overcoming the limitations of exploration with methods such as $\epsilon$-greedy policies \cite{watkins1989learning} or the introduction of noise in the actions of the agents \cite{DBLP:journals/corr/LillicrapHPHETS15} can be inefficient, especially in sparse long horizon tasks \cite{roman, 10.5555/3535850.3535978, 10.5555/3305890.3305968}. Even hand-crafting specific dense rewards necessitates substantial engineering efforts and limits generalisation as these are rendered highly case dependent \cite{10.5555/3305890.3305968, du2023guiding, amodei2016concrete}.

A promising alternative for encouraging exploration in such cases is the introduction of intrinsic rewards ($r^i$) \cite{10.5555/3535850.3535978, 10.5555/3305890.3305968, flet-berliac2021adversarially,burda2018exploration,oudeyer2007intrinsic,chentanez2005intrinsically}. While these methods enhance exploration in long-horizon, sparse-reward RL problems, these can still overemphasise noisy and non-relevant to the end-goal state transitions due to prediction errors \cite{10.5555/3305890.3305968, Taiga2020On}.

On the other hand, Large Language Models (LLMs) provide promising means of context and common-sense aware reasoning that could aid agents in otherwise complex settings, by emphasising more relevant state transitions based on the task at hand \cite{du2023guiding, yu2023language, radford2019language}. As such, the emergence of LLMs and their language-based instructions have shown promising applications in conjunction with deep learning \cite{du2023guiding, yu2023language, kwon2023reward, wu2023read, carta2023grounding} and in particular in the domain of robotics \cite{yu2023language, ahn2022i}. Nevertheless, while LLMs hold promise, it's essential to consider their limitations, as their outputs may not always be reliable nor optimal \cite{carta2023grounding, NEURIPS2020_1457c0d6}. Moreover, current methods utilising LLMs are faced with challenges, ranging from grounding and constraining LLMs \cite{carta2023grounding, ahn2022i, NEURIPS2020_1457c0d6}, limiting emergent behaviour by following specific manual guidelines \cite{wu2023read}, directly relying on goal suggestions that may deviate from the relevance of the task at hand \cite{du2023guiding} and often requiring significant prompt engineering efforts \cite{wei2023chainofthought}.

\begin{figure*}
    \centering
    \includegraphics[width=1\linewidth]{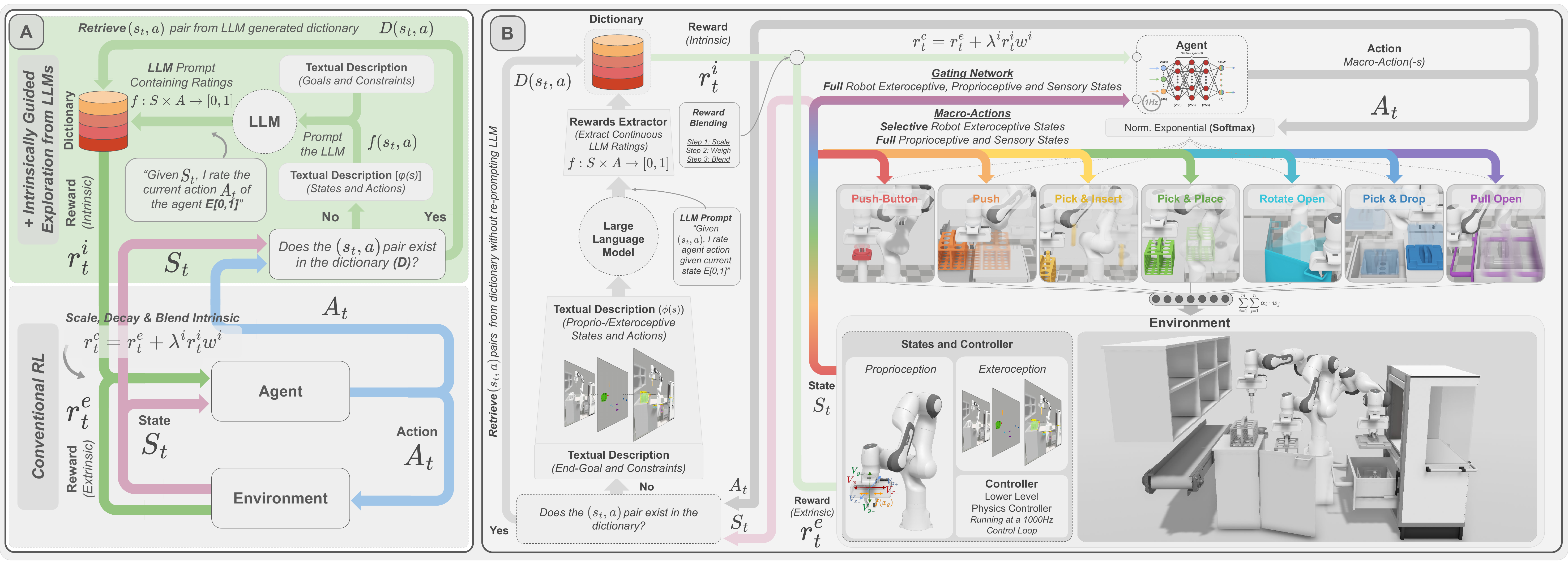}
    \caption{Schematics illustrating the principles of our method. \textbf{(A)} The overview of IGE-LLMs. \textbf{(B)} IGE-LLMs on ROMAN's hierarchical architecture \cite{roman} for solving complex robotic manipulation tasks entailing sparse rewards and long horizons.}
    \label{fig:method_visualised}
\end{figure*}

Instead, we propose IGE-LLMs (visualised in \autoref{fig:method_visualised}), a framework utilising LLMs as an assistive intrinsic proxy reward, alongside the traditional extrinsic in the conventional RL setting. In this way, we assert that even naturally occurring incorrect LLM replies \cite{carta2023grounding, NEURIPS2020_1457c0d6}, will not cause the policy to learn sub-optimally as the replies are used for guidance, with the extrinsic reward still eventually being the main policy driver. In particular, we hypothesise that utilising LLMs with their context-aware reasoning, can mitigate the limitations of current intrinsic methods exploring possibly non-related to the task states \cite{10.5555/3305890.3305968, Taiga2020On}. On the other hand, we assert that the integration of LLMs in conjunction with RL and the random exploration of the agent with its environment will compensate for the occasional inaccuracies of the LLM's outputs \cite{carta2023grounding, NEURIPS2020_1457c0d6}. We show that this synergy bridges the gaps of each approach, rather than using these in isolation. To the best of our knowledge, this is a current gap in the literature.

The contributions of our work are summarised as follows:
\begin{itemize}
\item A modular approach of utilising LLMs as intrinsic assistance, fostering exploration in complex environments with existing RL settings and algorithms; 
\item An extensive comparison of state-of-the-art intrinsic methods and related works on utilising LLMs;
\item A method capable of promoting exploration in complex, long-horizon sequential robotic manipulation tasks, entailing a plethora of macro-actions;
\item A computationally efficient approach of incorporating LLMs in RL paradigms whereby recurring situations of visited states are derived from a generated dictionary.
\end{itemize}

In the remainder, we present the related work, elaborate on the technical details of IGE-LLMs, and present the results comparing our method and state-of-the-art methods.

\section{RELATED WORK}
We consider related work that utilise LLMs, with a focus on the RL decision making process and work on intrinsically-motivated exploration, encompassing robot learning. 

\subsection{Reinforcement Learning -- Trials, Errors, and the Pursuit of Elusive Sparse Rewards}
Pre-programming robots as embodied intelligence via analytical models is sub-optimal due to the oversimplification of modelling real-world dynamics \cite{roman}. Deep learning algorithms \cite{LeCun2015}, particularly RL offer a promising alternative with their capacity to learn from interactions with the environment \cite{8793485, 8461249}, drawing inspiration from biology whereby even humans in their early lives must learn representations from unlabelled data \cite{Zaadnoordijk2022, Saxe2021}. The common deep RL algorithms are the Proximal Policy Optimization (PPO) \cite{schulman2017proximal} and Soft Actor-Critic (SAC) \cite{pmlr-v80-haarnoja18b}. While PPO is on-policy and generally less sample efficient than the off-policy SAC, we choose PPO in this work as it is less prone to instabilities and typically requires less hyperparameter tuning than SAC \cite{schulman2017proximal, 10.5555/3294996.3295141}.

In RL, agents aim to learn a set of policies by maximising their returns from the environment, known as extrinsic rewards. In general, when these extrinsic rewards are provided in an immediate or continuous manner, RL policies perform very well \cite{roman}. However, in many real-world robotic cases, these extrinsic rewards can be (i) significantly \textbf{sparse} exacerbating the exploration-exploitation trade-off \cite{Zaadnoordijk2022, 10.1145/3170427.3186500, 10.5555/3454287.3455660}, (ii) \textbf{miss-aligned} \cite{10.5555/3535850.3535978} or (iii) require very careful \textbf{crafting and tuning} which limits generalisation as these ultimately become task specific \cite{10.5555/3535850.3535978, du2023guiding}. Unfortunately, for agents to randomly reach a sparse goal, especially in long-horizon sequential robotic manipulation tasks \cite{davchev2021wish, roman}, is highly unlikely due to the increased need for exploration \cite{10.5555/3535850.3535978}, leading to resource-intensive learning sessions \cite{Thor2022, doi:10.1126/scirobotics.abb2174}.

\subsection{Intrinsic Guidance -- The Compass Navigating Through the Maze of Reinforcement Learning}
To tackle environments with sparse rewards, a prominent category of exploration techniques used are intrinsic rewards \cite{10.5555/3535850.3535978, 10.5555/3305890.3305968, burda2018exploration, Taiga2020On, bellemare2016unifying}. Intrinsic rewards ($r^i$) are computed and added to the extrinsic term ($r^e$), such that the total ($r^c$) is $r^c = r^e + \lambda r^i$, with $\lambda$ representing a scaling factor. Over time, $r^i$ is decayed, by which point the agent should converge towards the extrinsic reward, transitioning from exploration to exploitation \cite{10.5555/3535850.3535978}. There are two main categories, (i) \textbf{count-based} and (ii) \textbf{prediction-based} intrinsic reward methods.

Count-based exploration methods are traditional intrinsic methods that incentivise agents to visit rarely encountered states, useful in small, discrete state spaces \cite{bellemare2016unifying}. Nevertheless, these can be less effective in high-dimensional state spaces, commonly seen in complex robotic manipulation tasks entailing long horizon and sequential operation \cite{roman}. 

Recent prediction-based methods such as the Intrinsic Curiosity Module (ICM) \cite{10.5555/3305890.3305968} and the Random Network Distillation (RND) \cite{burda2018exploration} have been proposed. ICM utilises prediction error, promoting the agent to explore new and unfamiliar states, while RND generates intrinsic rewards by comparing feature representations of the agent with those of a fixed random network. Nonetheless, ICM can overemphasise noisy state transitions due to prediction error, potentially leading the agent to explore non-relevant to a given task state transitions \cite{10.5555/3305890.3305968, Taiga2020On}. On the other hand, RND's static random network might not always capture evolving complexities in dynamically changing environments such as physics-based interactions as with robotic manipulation \cite{roman, burda2018exploration, Taiga2020On}.

As it can be inferred, while intrinsic rewards can foster exploration, these may explore states beyond the relevance of the task at hand \cite{10.5555/3535850.3535978, 10.5555/3305890.3305968, burda2018exploration, Taiga2020On, bellemare2016unifying}. Perhaps, by merging the exploratory potential of intrinsic rewards in the form of contextual insights of LLMs, a more directed and efficient exploration strategy can be achieved.

\subsection{Guiding Reinforcement Learning with LLMs}
LLMs are capable of providing useful context as well as common-sense aware reasoning constituting them valuable in otherwise significantly complex environments \cite{du2023guiding, ahn2022i}. Nevertheless, a primary limitation of LLMs is that their outputs can occasionally be inaccurate \cite{carta2023grounding, NEURIPS2020_1457c0d6}. The work of \cite{wu2023read} utilised LLMs to extract information from game manuals via a QA extraction and reasoning module, whereby auxiliary rewards are inferred from "Yes/No" answers. While innovative, this inherently requires specific instructions, a reasoning module and ultimately an ideal behaviour is essentially hardwired by the content of the manual \cite{wu2023read}. 

The work of \cite{carta2023grounding}, introduced grounded LLMs serving as the agent's policy which is progressively updated via online RL. However, LLM's estimations cannot be guaranteed to be accurate \cite{NEURIPS2020_1457c0d6}, potentially leading to non-optimal behaviour. In another work, \cite{kwon2023reward} utilised LLMs to shape the behaviour of RL agents against user-specified objectives as textual prompts so as to align the agent's behaviour with human-described objectives. While novel, this framework is evaluated on rather short horizons, only provides binary reward signals and finally replaces the traditional reward function with a proxy reward rendering it reliant on the LLM's outputs which is prone to inaccuracies \cite{carta2023grounding, NEURIPS2020_1457c0d6}. The work of \cite{du2023guiding}, presented a goal-oriented approach whereby LLMs are used as the means of encouraging exploration in RL by rewarding the attainment of LLM-suggested goals based on and reliant on state descriptions to generate exploration goals.

Instead, in our method and unlike \cite{du2023guiding, kwon2023reward, wu2023read}, we utilise LLMs for the provision of a dense, continuous assistive intrinsic reward in environments naturally challenged by long-horizons and sparse extrinsic rewards. This intrinsic reward is decayed over time so as to not overemphasise plausible inaccuracies stemming from the LLM which can lead to bias \cite{carta2023grounding, NEURIPS2020_1457c0d6}. Instead, by decaying the intrinsic term, we assert that the agent will naturally converge to the extrinsic term, thus fostering emergent behaviour to achieve the end goal, while balancing exploitation and exploration. Moreover, our method facilitates learning efficiency in recurring situations whereby visited state-action pairs ($s_t, a_t$) can be derived from a generated dictionary, limiting the need for LLM inference.

\section{METHODOLOGY}
\subsection{Technical Preliminaries}
In this section, we outline technical preliminaries prior to the introduction of our method, including related (i) count-based and (ii) prediction-based intrinsic rewards.

\paragraph*{\textbf{Markov Decision Process}}
We consider a Markov Decision Process (MDP) as a tuple $(S, A, P, R, \gamma)$, whereby $S$ and $A$ represent the state and action spaces, respectively. $P(s'|s,a) = \Pr(s_{t+1}=s' | s_t=s, a_t=a)$ represents the transition function, specifying the probability of transitioning to the next state ($s'$) given the current state ($s$) and the applied action ($a$). The extrinsic reward function given as $R(s, a, s')$ provides the reward to the agent received after transitioning from state $s$ to state $s'$ given action $a$. The $\gamma \in [0,1]$ is the discount factor. The goal is to learn a policy $\pi(a|s)$ that defines the probability of taking action $a$ in state $s$ such that the expected discounted returns are maximised as $\mathbb{E}_\pi \left[ \sum_{t=0}^\infty \gamma^t \mathcal{R}(s_t, a_t, s_{t+1}) \mid a_t \sim \pi(s_t) \right]$. To promote exploration, we can also define a function that complements the extrinsic reward function $R$ with an intrinsic reward, which allows the agent to learn to explore interesting states.

\paragraph*{\textbf{Count-Based Intrinsic Reward}} With count-based intrinsic rewards, the agents are encouraged to visit states that have not frequently been encountered \cite{10.5555/3535850.3535978, bellemare2016unifying, ostrovski2017count}. These rewards are inverse proportional to the visitations of already encountered states and are commonly represented as: $r_t^i = \frac{1}{\sqrt{N(s_t)}}$, whereby $r_t^i$ denotes the intrinsic reward at time-step $t$ and $N(s_t)$ the count of the encountered current state. Thereby, higher intrinsic rewards are given for states that have been visited less frequently, encouraging exploration.

\paragraph*{\textbf{Intrinsic Curiosity Module (ICM)}} With ICM, an intrinsic reward signal is introduced based on the ability of the agent to predict the consequence of its actions in a learned feature space. The intrinsic reward is formulated as $r^i_t = \alpha( \widehat{\phi}(s_{t+1}) - \phi(s_{t+1}))^2$, whereby $\alpha$ is a scaling factor. $\widehat{\phi}$ denotes the predicted representation of the next state given the current state and action, while $\phi$ represents the actual representation of the next state. In this way, the model attempts to learn to encode information affected by the action of the agent. Essentially, $r^i$ represents the prediction error of the agent's estimate of the next state's feature representation. Thus, the agent is encouraged to perform actions that maximise the error, exploring novel areas \cite{10.5555/3305890.3305968}.

\paragraph*{\textbf{Random Network Distillation (RND)}} RND introduces a prediction-based intrinsic reward signal that is derived from the ability of the learned network, referred to as the predictor, to mimic a randomly initialised network, referred to as the target network. The intrinsic reward $r^i$ is given as $r^i_t = ( \widehat{\phi}(s_{t + 1}) - \phi(s_{t + 1}))^2$, whereby $\widehat{\phi}$ and $\phi$ represent the predictor and target networks respectively for state $s$ \cite{burda2018exploration}.

\subsection{Intrinsically Guided Exploration from LLMs (IGE-LLMs)}
We propose a novel way of utilising LLMs to encourage efficient exploration in RL, particularly in long-horizon sequential tasks with sparse rewards. In our approach, given a state $s_t$ of the environment at a given time $t$, the LLM is tasked with evaluating the potential future rewards of all actions $a \in A$ the agent is capable of performing. This evaluation is performed by a function $f: S \times A \rightarrow [0,1]$, whereby $S$ and $A$ represent the state and action spaces respectively. The output of the action evaluation is represented as a continuous scalar value in $[0,1]$. For every pair of states and actions $(s_t, a)$, the LLM is tasked with assigning a rating $r^i = f(s_t, a)$, representing the desirability of performing action $a$ at state $s_t$. Intuitively, $f$ guides the agent towards actions that the LLM perceives as most beneficial given $s_t$.

This process can be repeated once for every new state, prompting the LLM to rate the available actions the agent can perform and build a dictionary for every seed. By creating a dictionary, $D: S \times A \rightarrow [0,1]$, computational costs of running inference on the LLM or Application Programming Interface (API)-related costs can be kept to a minimum. In this process, whenever the agent encounters a state $s_t$, it retrieves the corresponding intrinsic rewards for all actions from the dictionary $D$. If the state-action pair $(s_t, a)$ already exists in $D$ and hence previously encountered, the intrinsic reward is directly obtained as $r^i = D(s_t, a)$. Alternatively, if the state-action pair is not present in $D$, hence not yet encountered, the LLM is prompted to compute the intrinsic reward $r^i = f(s_t, a)$, which is subsequently stored as $D(s_t, a) = r^i$. In this way, the dictionary serves as an efficient guide for future actions of the agent in similar states.

By leveraging the existing RL paradigm, the total reward is given as: $r^c = r^e + \lambda^i \cdot r^i \cdot w^i$, representing the combined reward ($r^c$), entailing the sum of the extrinsic ($r^e$) and intrinsic ($r^i$) terms. The intrinsic term is furthermore controlled by a scaling factor ($\lambda^i$) as well as a linearly decaying weight ($w^i$) to guide the policy, without overfitting it to the intrinsic reward, due to naturally occurring inaccuracies stemming from LLMs \cite{NEURIPS2020_1457c0d6}. In this way, the agent receives a decaying over-time dense intrinsic reward and gradually learns to converge towards the sparse extrinsic from the environment. We hypothesise this approach will encourage exploration in sparse reward environments entailing long horizons, commonly seen in complex robotic manipulation tasks \cite{roman}. \autoref{fig:method_visualised} visually depicts our method -- IGE-LLMs.

\subsection{Apparatus and System Configuration}
Combining learning algorithms with simulators mimicking real-world physics offers notable advantages \cite{8793485, doi:10.1126/scirobotics.abb2174}, facilitating faster robotic learning and minimising the risks of hardware damage \cite{8793485, doi:10.1126/scirobotics.abb2174, 9492850}. Hence, to achieve a realistic learning-based robotic simulation as with \cite{roman}, we utilised the simulation engine Unity3D, incorporating NVIDIA's physics engine and the PyTorch-based ML-Agents toolkit for learning \cite{juliani2018unity}. Moreover, the Robotics Operating System (ROS) ROS\# plugin ensured accurate physical modelling by importing the Franka Emika robot's characteristics (Unified Robot Description Format). The physics frequency was set to $1000$Hz to ensure stable performance \cite{doi:10.1126/scirobotics.abb2174, 9492850}. We employed the GPT-4 LLM, without any fine-tuning.

To emulate a realistic robotic setup, we also utilise a vision system similar to \cite{roman}, to detect the exteroceptive states and in particular the Objects of Interest (OIs) from the scene. The OIs are visualised in colour in \autoref{fig:method_visualised}.B as well as \autoref{fig:ige_llms_environments}.B. The system implements an object detection module based on the VGG-16 backbone \cite{8202133}, initialised with pre-trained weights on the ImageNet dataset and fine-tuned with the OIs in the simulation. This system generated textual scene descriptions for the LLM, indicating the presence or absence of OIs.

\section{Extending the RObotic MAnipulation Network (ROMAN)}
In this work, we validate our proposed method and related methods on a complex, long-horizon sequential robotic manipulation task, as depicted in \autoref{fig:method_visualised}.B, \autoref{fig:ige_llms_environments}.B and  \autoref{fig:ige_llms_on_roman} and based on \cite{roman}. We hypothesise that LLMs can be of significant benefit in hierarchical architectures especially those tasked with completing complex long-horizon robotic manipulation tasks necessitating the correct sequential coordination and activation of a plethora of different macro-actions \cite{roman}. Moreover, in such settings, the completion of a long-horizon sequential goal is contingent upon the successful completion of other sub-tasks; for instance, one cannot retrieve an item from a drawer without first opening it. To test our hypothesis, we extend the work known as ROMAN \cite{roman}, an event-based Hybrid Hierarchical Learning (HHL) architecture composed of distinct specialising experts treated hereinafter as \textbf{macro-actions}, commonly seen in real-life \cite{Billardeaat8414}. 

\subsection{Main Environment}
We concentrate on the environment where ROMAN was assessed on, see \autoref{fig:method_visualised}.B and \autoref{fig:ige_llms_environments}.B. The environment consisted of a laboratory setting, with the primary objective of retrieving a vial, placing it in a rack, and moving it onto a conveyor belt. Within this setting, additional sub-tasks were derived, ensuring their interdependence and relevance \cite{roman, Billardeaat8414}. We maintain the macro-actions and evaluate the methodologies of this study hereinafter on the gating network's ability to coordinate these to solve intricate long-horizon sequential robotic manipulation tasks. We hypothesise that our method, applied to ROMAN's gating network, will leverage the LLM's context-aware reasoning for enhanced exploration.

\subsection{Internal Hybrid Learning Procedure}
The expert Neural Network (NN)s, treated as macro-actions, were trained in a hybrid manner, combining Behavioural Cloning (BC) \cite{10.5555/3304652.3304697}, RL (PPO) \cite{schulman2017proximal}, and Generative Adversarial Imitation Learning (GAIL) \cite{NIPS2016_cc7e2b87}. Initially, policies were warm-started with BC, thereafter updated using RL (PPO) with extrinsic ($r^e$) and intrinsic ($r^i$) rewards from the environment and the GAIL discriminator respectively. Each NN received $N=20$ demonstrations from keybindings, corresponding to velocity controlling the end-effector and its binary gripper state. For further details consult \cite{roman}.

\subsection{Expert NNs - States, Actions and Rewards} 
The proprioceptive states were identical for all experts. Exteroceptive states differed, dependent on each NN's specialising manipulation skill. This allowed each NN to focus on its own core exteroceptive information relevant to its sub-task, influenced from a neuro-scientific perspective whereby humans determine information relevance during motor tasks \cite{Wolpert2011}. All experts shared identical actions, controlling the velocity of the robotic end-effector and the binary gripper state. A sparse, terminal reward ($+1$) was given upon completion of their specific sub-task goal. For more details consult \cite{roman}.

\begin{figure*}
    \centering
    \includegraphics[width=1.0\linewidth]{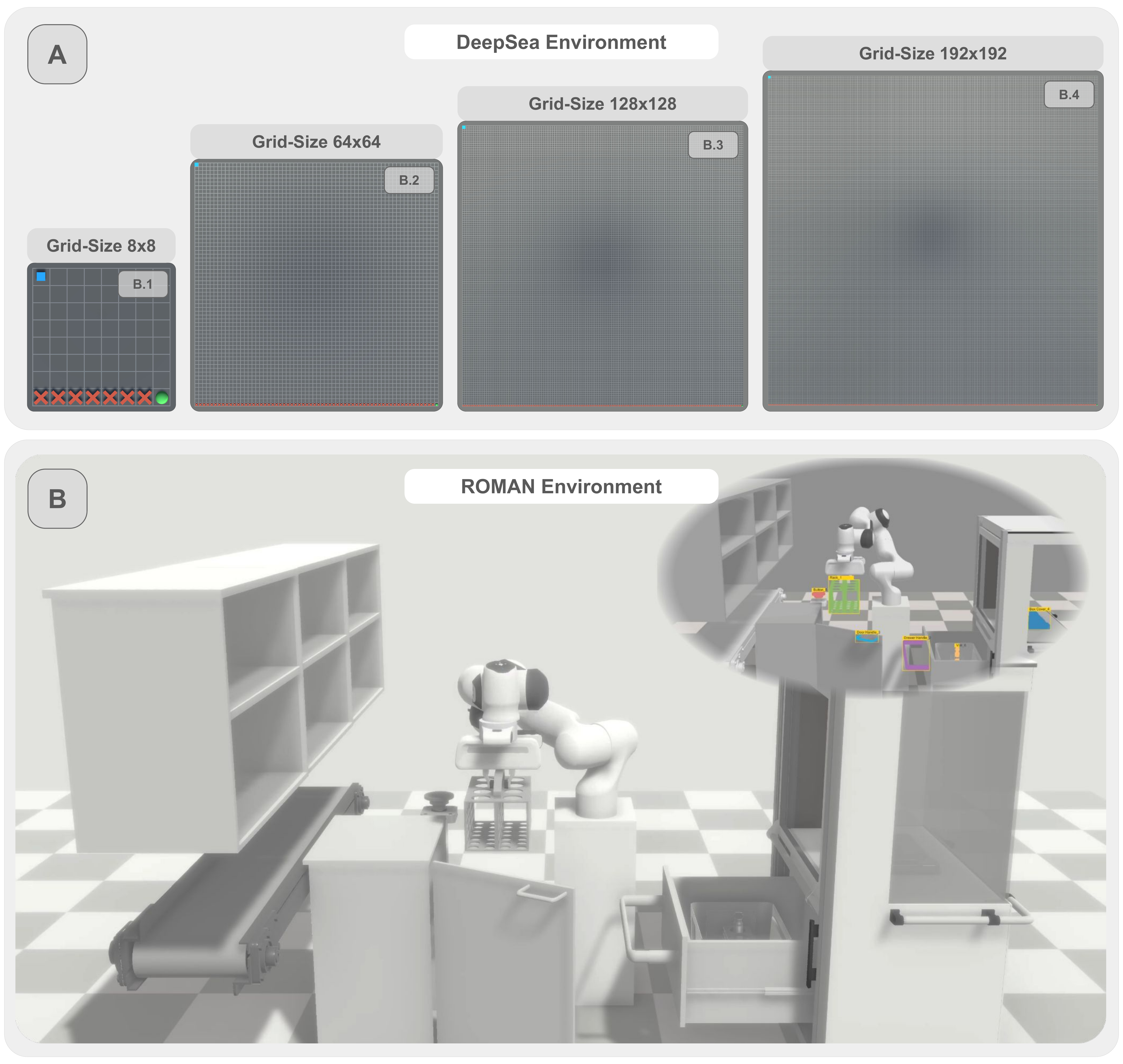}
    \caption{The two environments employed for evaluating all methods in the chapter including the proposed IGE-LLMs framework. \textbf{Figure A:} The preliminary grid-based environment -- DeepSea. From left to right, increasing grid sizes of the DeepSea environment, notably increasing the dimensionality problem and by extent the degrees of exploration necessitated to achieve the goal. Colours of blue, green and red represent the agent, goal and traps respectively. \textbf{B.1-B.4} represent grid sizes \textit{8x8}, \textit{64x64}, \textit{128x128} and \textit{192x192} respectively. DeepSea is challenged by exploration. Note on Visualisation for DeepSea: For the grid sizes \textit{128x128} and \textit{192x192}, the agent virtual mesh size, depicted in blue colour, is increased four-fold for the purpose of enhanced visualisation. \textbf{Figure B:} The main environment, ROMAN, studying notably intricate robotic manipulation tasks, necessitating the correct sequential orchestration of a plethora of macro-actions. It is worthwhile to point out that all related methods, including IGE-LLMs, were applied to ROMAN's gating network directly. Consequently, the evaluation for the main robotic environment entailed how well the gating network is capable of inferring and orchestrating the different macro-actions in a robotic manipulation task challenged by both exploration and long-horizons.}
    \label{fig:ige_llms_environments}
\end{figure*}

\section{EVALUATION}
We evaluate our method against the state-of-the-art in two environments (depicted in \autoref{fig:ige_llms_environments}): a toy setting targeting the challenge of RL exploration and a main environment entailing an intricate sequential robotic manipulation task challenged by both exploration and long-horizons. Each training seed utilising the LLM, corresponds to a different generated dictionary, ensuring randomness and diverse prompts. The prompts used for input to the LLM can be found in \autoref{fig:ige_llms_prompts}. Lastly, a function $\phi(s)$ is employed to convert state vectors into textual descriptions for LLM input, given direct state vectors might be uninterpretable for LLMs \cite{yu2023language}. For consistency and fairness with baselines, high-level descriptions are also integrated as binary values in the state vector.

\subsection{Illustrating the idea on a Toy Environment -- DeepSea}
Prior to proceeding to the main robotic task, we first experiment in a $N$x$N$ grid-based environment -- DeepSea, depicted in \autoref{fig:ige_llms_environments}.A. DeepSea is proposed as part of the Behaviour Suite (Bsuite) for RL whereby the challenge of exploration is targeted \cite{Osband2020Behaviour}. The agent starts at the top-left and aims to reach the goal located in the bottom right grid. We evaluate increasingly more complex grid sizes of $N \in \{8, 64, 128, 192\}$.

\paragraph{States, Actions and Rewards} The state space ($s_t$) included the agent's position ($x1, y1$) and target goal ($x2, y2$). Actions ($a_t$) corresponded to moving down ($x, y+1$) or diagonally down-right ($x+1, y+1$). The goal yielded a +1 sparse reward; while other bottom tiles incurred a -1 penalty.

\paragraph{LLM Implementation} The LLM is provided a prompt at each time-step of the DeepSea environment. Similarly to the NNs having access to the positions of the agent and the goal, the LLM is provided identical information in the form of textual descriptions. The LLM is tasked to rate $E(0,1)$ the two possible actions ($a_t$) of the agent, given the current position of the agent and the goal's constant position. Consult \autoref{fig:ige_llms_prompts}.1.A for the prompts utilised for the LLM.

\subsection{Complex Long-Horizon of Sparse Rewards Robotic Tasks}
For the main environment, we study a set of complex robotic manipulation tasks (see \autoref{fig:method_visualised}.B), commonly seen in robotics and physics-based interactions \cite{roman, Billardeaat8414}.

\paragraph{States, Actions and Rewards} 
In this environment, depicted in \autoref{fig:ige_llms_environments}.B, the gating network was the primary NN used for the evaluation and its ability to orchestrate its macro-actions to achieve the long-horizon end goal. We incorporate a softmax function, $\sigma (\mathbf {z} )_{i}={\frac {e^{z_{i}}}{\sum _{j=1}^{K}e^{z_{j}}}}$ to normalise the sum of weights of the seven macro-actions. The state space comprised the combined exteroceptive states of all its macro-actions in the hierarchy, allowing it to oversee the entirety of the environment. Its proprioceptive states included the end-effector's position, velocity and gripper state. A sparse terminal $+1$ reward was provided only upon completing the sequential end-goal. Given the high complexity of the task, we posit that dense intrinsic rewards by the LLM with their context-aware reasoning can aid learning in long-horizon manipulation.

\begin{figure*}
    \centering
    \includegraphics[width=1.0\linewidth]{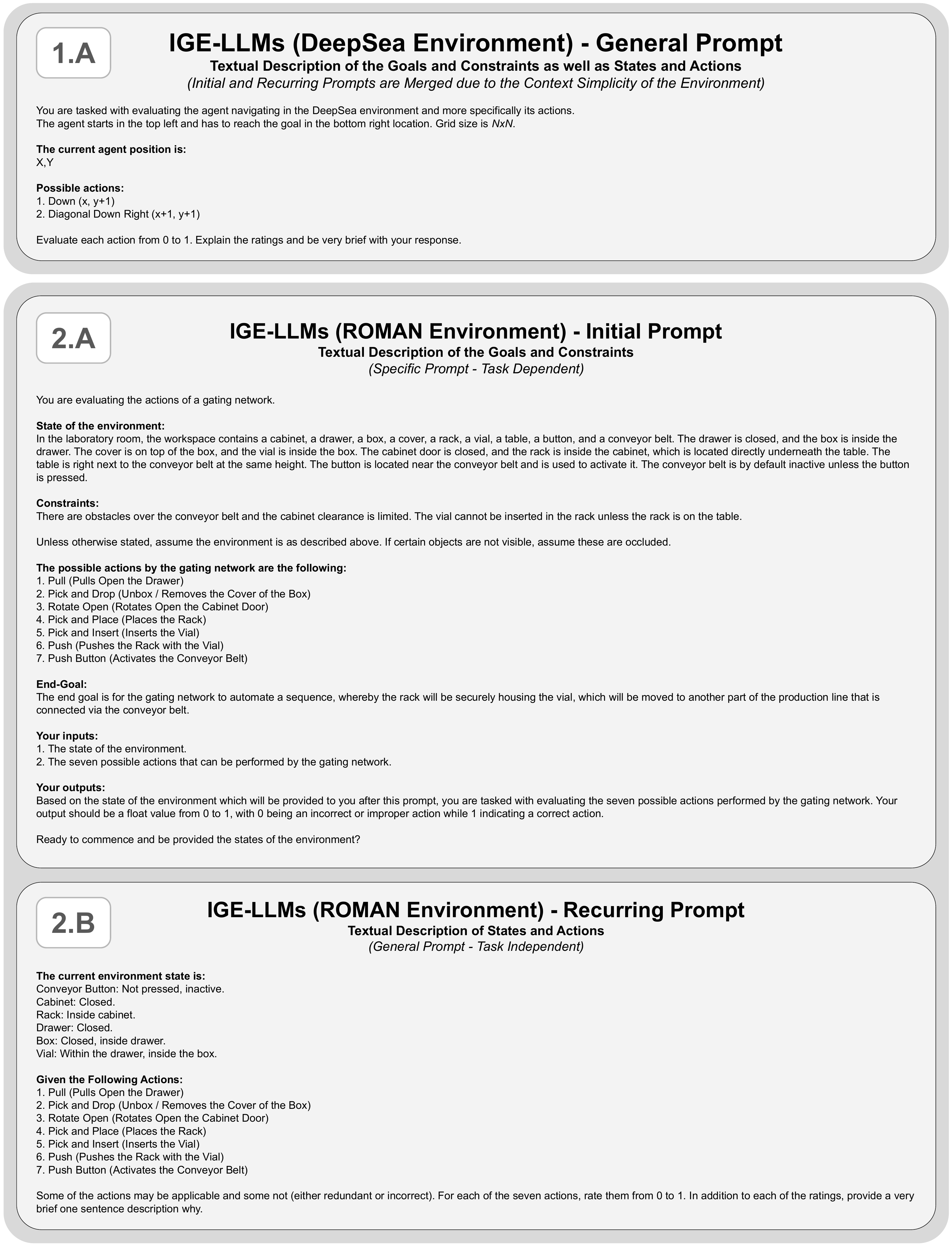}
    \caption{The textual prompts. \textbf{Figure 1.A:} The textual prompt description provided as input to the LLM for the DeepSea environment. \textbf{Figure 2:} The textual prompt descriptions provided as input to the LLM for the ROMAN environment. \textbf{Figure 2.A:} The \textit{initial prompt} entailing the goal and constraints of the environment. \textbf{Figure 2.B:} The \textit{recurring prompt} outlining the states and actions of the agent, whereby the LLM is tasked to compute the intrinsic reward ($r^i \in [0, 1]$).}
    \label{fig:ige_llms_prompts}
\end{figure*}

\paragraph{LLM Implementation} The LLM is provided a prompt at each time-step equivalent to the frequency of the gating NN (\(1Hz\)). The LLM is provided high-level textual descriptions corresponding to the binary states of the primary OIs in the scene, which is also integrated as binary values in the state vector to ensure fairness with baseline comparisons. At each time-step the LLM is asked to rate $E(0,1)$ the seven possible macro actions ($a_t$), based on the textual description of the OIs, rating as such the state-action pairs ($s_t, a$). Consult \autoref{fig:ige_llms_prompts}.2.A-B for the prompts utilised for the LLM.

\begin{figure*}
    \centering
    \includegraphics[width=1.0\linewidth]{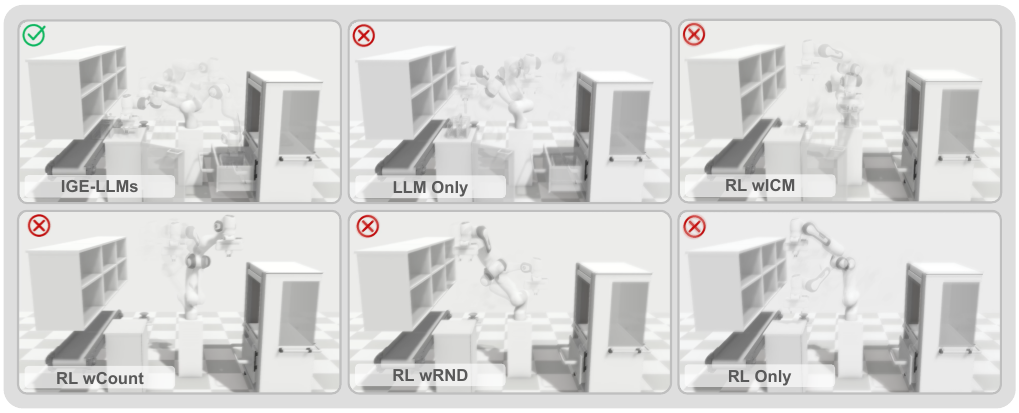}
    \caption{IGE-LLMs on ROMAN's longest-horizon task, case seven, at $\sigma = \pm 0.5$cm noise, compared against other methods. For more details of the sequential failures exhibited by other methods please consult the video available at \cite{Triantafyllidis2024Intrinsic}.}
    \label{fig:ige_llms_on_roman}
\end{figure*}

\section{RESULTS} 
\begin{figure*}
    \centering
    \includegraphics[width=1\linewidth]{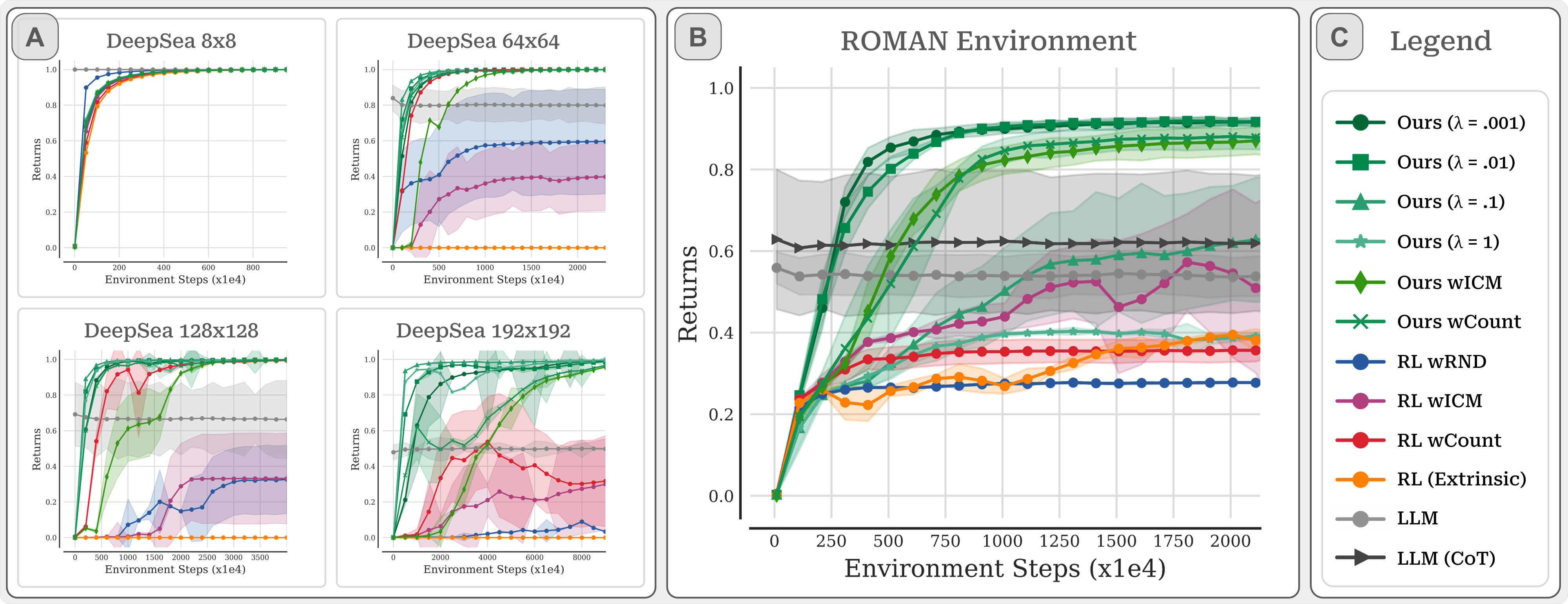}
    \caption{Normalised evaluation returns. Shading depicts the standard deviation ($\sigma$) around the mean. \textbf{A} DeepSea, from $n=5$ seeds for \textit{8x8} and \textit{64x64} and $n=3$ seeds for \textit{128x128} and \textit{192x192}. \textbf{B} ROMAN from $n=5$ seeds. \textbf{C} Legend.}
    \label{fig:results_returns}
\end{figure*}

We first compare conventional RL utilising extrinsic rewards (\textbf{RL}), thereafter combining it with intrinsic methods, including prediction-based (\textbf{ICM}, \textbf{RND}), \textbf{count-based}, and our \textbf{IGE-LLM}. Next, we combine our method with existing intrinsic methods to demonstrate its modularity and robustness. Additionally, to underscore the inherent limitations of occasional prompt inaccuracies stemming from LLMs \cite{carta2023grounding, NEURIPS2020_1457c0d6}, we also compare the LLM's direct output, without training. To this end, we also utilise state-of-the-art reasoning referred to as Chain of Thought (CoT) \cite{wei2023chainofthought}, due to its ability to elicit significantly improved arithmetic, commonsense, and symbolic reasoning. For both direct \textbf{LLM} and \textbf{LLM with CoT} outputs, we use the \textbf{argmax}. To incorporate randomness into the decision-making process with the argmax, when identically rated actions by the LLM are present in the dictionary, a random selection is made amongst those occurrences. Lastly, we perform a sensitivity analysis of our method and a robustness test of all trained models compared in this work. Smoothing and averaging are applied to \autoref{fig:results_returns} for visualisation purposes. Refer to the accompanying video for demonstrations of the models in the environments \cite{Triantafyllidis2024Intrinsic}.

\begin{figure*}
    \centering
    \includegraphics[width=\linewidth]{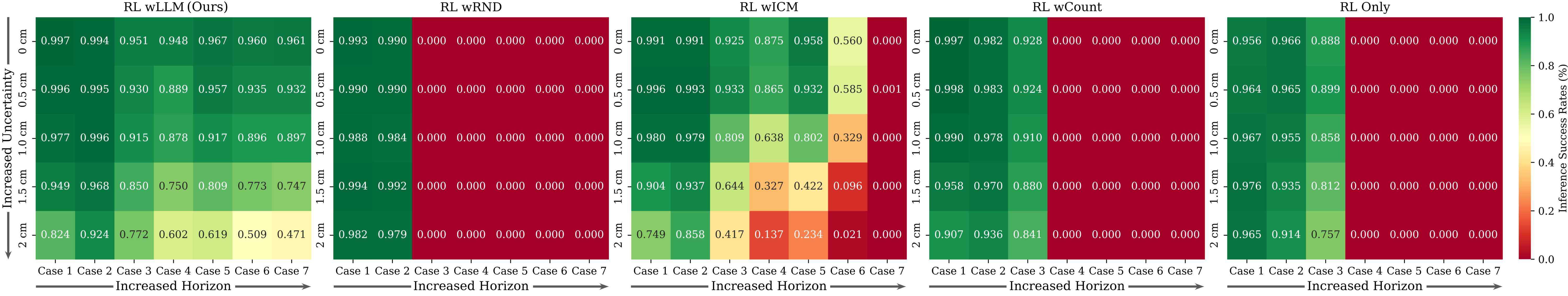}
    \caption{Inference results for the ROMAN environment across five distinct models. The x-axis represents the task horizon, ascending from left to right, y-axis the exteroceptive noise, ascending from top to bottom. Each cell stems from 1K episodes.}
    \label{fig:inference_roman}
\end{figure*}

\subsection{Using LLMs as Intrinsic Assistance}
Results for DeepSea and the main ROMAN environments are depicted in \autoref{fig:results_returns}. It is inferred that while existing intrinsic reward methods ($r^i$) are overall beneficial in promoting exploration in sparse environments compared to just the use of RL ($r^e$), these still struggle with increased dimensionality and horizons. Notably, RL+Count performs mostly well for DeepSea, yet shows significant drops in returns in ROMAN. RL+RND and RL+ICM both struggle with increased grid sizes in DeepSea and ROMAN, yet RL+ICM shows higher returns in ROMAN, still not close to the maximum attainable. Conversely, IGE-LLMs consistently performs well in DeepSea across all tested grid sizes. Most importantly, IGE-LLMs outperforms by a notable margin all other methods in ROMAN. Moreover, \autoref{fig:results_returns} shows that IGE-LLMs can also be combined and complements existing intrinsic methods, highlighting its modularity and robustness.

\subsection{The Constraints of Direct Integration of LLMs}
The direct integration of LLMs into the decision-making, as seen in \autoref{fig:results_returns}, exhibit significant errors, even with advanced LLMs (\textit{GPT-4}) prompting occasional inaccuracies. The argmax results in \autoref{fig:results_returns} show that the direct integration of the LLM, even when incorporating CoT \cite{wei2023chainofthought} in ROMAN, is inadequate, especially for complex long-horizon robotic manipulation tasks. It is concluded that directly relying on LLMs should be avoided, in line with \cite{carta2023grounding, NEURIPS2020_1457c0d6}. Instead, results show that utilising LLMs as an assistive intrinsic signal is preferable. In this way, the agent's innate interactions within its environment, as per the RL paradigm, can counteract the inherent limitations when directly relying on and integrating LLMs in the decision-making process.

\subsection{Sensitivity Analysis}
A common challenge when employing intrinsic-motivated exploration, is the sensitivity to the intrinsic scale $\lambda$ \cite{10.5555/3535850.3535978}. Hence, to evaluate the sensitivity of our method, we evaluate different logarithmic spaced scaling parameters $\lambda^i \in \{0.001, 0.01, 0.1, 1\}$. In this way, the robustness of our approach can be demonstrated from a fraction to exceeding $\lambda$ values to the extrinsic reward. From \autoref{fig:results_returns} results show that our method exhibits robustness to varying scaling parameters, suggesting that the use of LLMs as intrinsic assistance is fairly insensitive to changes in $\lambda$. However, some influence is observed with those scales equaling and exceeding $r^e$ and in particular $\lambda^i \in \{1\}$ for \autoref{fig:results_returns}.B. These results suggest that overemphasis on the LLM's output is discouraged due to its occasional inaccuracies \cite{carta2023grounding, NEURIPS2020_1457c0d6}, as evidenced by the LLM and LLM CoT results.

\subsection{Inference Results - Robustness Test}
To further evaluate our model's robustness, we add Gaussian noise to all exteroceptive states in ROMAN's gating NN, linked to positional observations. We test from 0cm to up to $\sigma = \pm 2 \text{cm}$ noise levels, in $0.5\text{cm}$ increments. From \autoref{fig:inference_roman}, it is inferred that utilising $r^i$ is overall beneficial compared to using just $r^e$. In particular, while RL wICM shows higher success rates than other methods, errors increase notably with longer horizons and higher uncertainty. RL, RL wRND and RL wCount appear to struggle with longer horizons, showing that Cases 4 to 7 are effectively unattainable. In contrast, IGE-LLMs, maintains high success rates across all uncertainty levels and horizons, outperforming RL and other intrinsic methods. \autoref{fig:ige_llms_on_roman} depicts the inference results.

\section{Discussion and Conclusion}
\subsection{Limitations and Future Work}
While most methods utilising LLMs require textual scene descriptions, methods such as Contrastive Language-Image Pre-Training (CLIP) \cite{radford2021learning} could automate the description of exteroceptive states. Moreover, combining our method with \cite{yu2023language} to bridge high-level instructions to robotic actions via LLMs, investigating different levels of weight decays ($w^i$) alongside the scaling parameters ($\lambda^i$) and a comparison between different LLM models, may prove to be valuable extensions. Nevertheless, our results showed that even a very capable LLM model (\textit{gpt-4}) was overall inadequate to be used in isolation, highlighting the value of using LLMs as assistive guidance rather than directly as main policy drivers.

\subsection{Conclusion}
This work presented IGE-LLMs, a novel approach leveraging LLMs for intrinsic assistance, promoting exploration in RL tasks challenged by sparse rewards and long-horizons. Validated on a preliminary and subsequent intricate long-horizon robotic manipulation environment, results showed IGE-LLMs outperformed existing intrinsic methods and underlined the shortcomings of the direct use of LLMs in the decision making process. Furthermore, IGE-LLMs modularity was underscored by its ability to be combined and complement existing intrinsic methods, its insensitivity to most scaling parameters featured, and its consistent robustness against increased uncertainty and horizons highlighted. Collectively, these findings underline the value of IGE-LLMs for tackling the exploration and long-horizon challenges inherent in complex robotic manipulation tasks.

\section*{ACKNOWLEDGMENT}
Supported by the EPSRC CDT in RAS (EP/L016834/1).

\balance
\bibliographystyle{IEEEtran}
\bibliography{bibliography}

\begin{thebibliography}{10}
\providecommand{\url}[1]{#1}
\csname url@rmstyle\endcsname
\providecommand{\newblock}{\relax}
\providecommand{\bibinfo}[2]{#2}
\providecommand\BIBentrySTDinterwordspacing{\spaceskip=0pt\relax}
\providecommand\BIBentryALTinterwordstretchfactor{4}
\providecommand\BIBentryALTinterwordspacing{\spaceskip=\fontdimen2\font plus
\BIBentryALTinterwordstretchfactor\fontdimen3\font minus \fontdimen4\font\relax}
\providecommand\BIBforeignlanguage[2]{{%
\expandafter\ifx\csname l@#1\endcsname\relax
\typeout{** WARNING: IEEEtran.bst: No hyphenation pattern has been}%
\typeout{** loaded for the language `#1'. Using the pattern for}%
\typeout{** the default language instead.}%
\else
\language=\csname l@#1\endcsname
\fi
#2}}

\bibitem{davchev2021wish}
\BIBentryALTinterwordspacing
T.~Davchev, O.~O. Sushkov, J.-B. Regli, S.~Schaal, Y.~Aytar, M.~Wulfmeier, and J.~Scholz, ``Wish you were here: Hindsight goal selection for long-horizon dexterous manipulation,'' in \emph{International Conference on Learning Representations}, 2022. [Online]. Available: \url{https://openreview.net/forum?id=FKp8-pIRo3y}
\BIBentrySTDinterwordspacing

\bibitem{8843293}
R.~Fox, R.~Berenstein, I.~Stoica, and K.~Goldberg, ``Multi-task hierarchical imitation learning for home automation,'' in \emph{2019 IEEE 15th International Conference on Automation Science and Engineering (CASE)}, 2019, pp. 1--8.

\bibitem{roman}
\BIBentryALTinterwordspacing
E.~Triantafyllidis, F.~Acero, Z.~Liu, and Z.~Li, ``Hybrid hierarchical learning for solving complex sequential tasks using the robotic manipulation network {ROMAN},'' \emph{Nature Machine Intelligence}, vol.~5, no.~9, pp. 991--1005, Sep 2023. [Online]. Available: \url{https://doi.org/10.1038/s42256-023-00709-2}
\BIBentrySTDinterwordspacing

\bibitem{10.5555/3535850.3535978}
L.~Sch\"{a}fer, F.~Christianos, J.~P. Hanna, and S.~V. Albrecht, ``Decoupled reinforcement learning to stabilise intrinsically-motivated exploration,'' in \emph{Proceedings of the 21st International Conference on Autonomous Agents and Multiagent Systems}, ser. AAMAS '22.\hskip 1em plus 0.5em minus 0.4em\relax Richland, SC: International Foundation for Autonomous Agents and Multiagent Systems, 2022, p. 1146–1154.

\bibitem{watkins1989learning}
C.~J. C.~H. Watkins, ``Learning from delayed rewards,'' Ph.D. dissertation, King's College, Cambridge, 1989.

\bibitem{DBLP:journals/corr/LillicrapHPHETS15}
\BIBentryALTinterwordspacing
T.~P. Lillicrap, J.~J. Hunt, A.~Pritzel, N.~Heess, T.~Erez, Y.~Tassa, D.~Silver, and D.~Wierstra, ``Continuous control with deep reinforcement learning,'' in \emph{4th International Conference on Learning Representations, {ICLR} 2016, San Juan, Puerto Rico, May 2-4, 2016, Conference Track Proceedings}, Y.~Bengio and Y.~LeCun, Eds., 2016. [Online]. Available: \url{http://arxiv.org/abs/1509.02971}
\BIBentrySTDinterwordspacing

\bibitem{10.5555/3305890.3305968}
D.~Pathak, P.~Agrawal, A.~A. Efros, and T.~Darrell, ``Curiosity-driven exploration by self-supervised prediction,'' in \emph{Proceedings of the 34th International Conference on Machine Learning - Volume 70}, ser. ICML'17.\hskip 1em plus 0.5em minus 0.4em\relax JMLR.org, 2017, p. 2778–2787.

\bibitem{du2023guiding}
Y.~Du, O.~Watkins, Z.~Wang, C.~Colas, T.~Darrell, P.~Abbeel, A.~Gupta, and J.~Andreas, ``Guiding pretraining in reinforcement learning with large language models,'' 2023.

\bibitem{amodei2016concrete}
D.~Amodei, C.~Olah, J.~Steinhardt, P.~Christiano, J.~Schulman, and D.~Mané, ``Concrete problems in ai safety,'' 2016.

\bibitem{flet-berliac2021adversarially}
\BIBentryALTinterwordspacing
Y.~Flet-Berliac, J.~Ferret, O.~Pietquin, P.~Preux, and M.~Geist, ``Adversarially guided actor-critic,'' in \emph{International Conference on Learning Representations}, 2021. [Online]. Available: \url{https://openreview.net/forum?id=_mQp5cr_iNy}
\BIBentrySTDinterwordspacing

\bibitem{burda2018exploration}
\BIBentryALTinterwordspacing
Y.~Burda, H.~Edwards, A.~Storkey, and O.~Klimov, ``Exploration by random network distillation,'' in \emph{International Conference on Learning Representations}, 2019. [Online]. Available: \url{https://openreview.net/forum?id=H1lJJnR5Ym}
\BIBentrySTDinterwordspacing

\bibitem{oudeyer2007intrinsic}
P.-Y. Oudeyer, F.~Kaplan, and V.~V. Hafner, ``Intrinsic motivation systems for autonomous mental development,'' \emph{IEEE Transactions on Evolutionary Computation}, vol.~11, no.~2, pp. 265--286, 2007.

\bibitem{chentanez2005intrinsically}
\BIBentryALTinterwordspacing
N.~Chentanez, A.~Barto, and S.~Singh, ``Intrinsically motivated reinforcement learning,'' in \emph{Advances in Neural Information Processing Systems}, L.~Saul, Y.~Weiss, and L.~Bottou, Eds., vol.~17.\hskip 1em plus 0.5em minus 0.4em\relax MIT Press, 2004. [Online]. Available: \url{https://proceedings.neurips.cc/paper_files/paper/2004/file/4be5a36cbaca8ab9d2066debfe4e65c1-Paper.pdf}
\BIBentrySTDinterwordspacing

\bibitem{Taiga2020On}
\BIBentryALTinterwordspacing
A.~A. Taiga, W.~Fedus, M.~C. Machado, A.~Courville, and M.~G. Bellemare, ``On bonus based exploration methods in the arcade learning environment,'' in \emph{International Conference on Learning Representations}, 2020. [Online]. Available: \url{https://openreview.net/forum?id=BJewlyStDr}
\BIBentrySTDinterwordspacing

\bibitem{yu2023language}
W.~Yu, N.~Gileadi, C.~Fu, S.~Kirmani, K.-H. Lee, M.~G. Arenas, H.-T.~L. Chiang, T.~Erez, L.~Hasenclever, J.~Humplik, B.~Ichter, T.~Xiao, P.~Xu, A.~Zeng, T.~Zhang, N.~Heess, D.~Sadigh, J.~Tan, Y.~Tassa, and F.~Xia, ``Language to rewards for robotic skill synthesis,'' 2023.

\bibitem{radford2019language}
A.~Radford, J.~Wu, R.~Child, D.~Luan, D.~Amodei, I.~Sutskever, \emph{et~al.}, ``Language models are unsupervised multitask learners,'' \emph{OpenAI blog}, vol.~1, no.~8, p.~9, 2019.

\bibitem{kwon2023reward}
M.~Kwon, S.~M. Xie, K.~Bullard, and D.~Sadigh, ``Reward design with language models,'' 2023.

\bibitem{wu2023read}
Y.~Wu, Y.~Fan, P.~P. Liang, A.~Azaria, Y.~Li, and T.~M. Mitchell, ``Read and reap the rewards: Learning to play atari with the help of instruction manuals,'' 2023.

\bibitem{carta2023grounding}
T.~Carta, C.~Romac, T.~Wolf, S.~Lamprier, O.~Sigaud, and P.-Y. Oudeyer, ``Grounding large language models in interactive environments with online reinforcement learning,'' 2023.

\bibitem{ahn2022i}
M.~Ahn, A.~Brohan, N.~Brown, Y.~Chebotar, O.~Cortes, B.~David, C.~Finn, C.~Fu, K.~Gopalakrishnan, K.~Hausman, A.~Herzog, D.~Ho, J.~Hsu, J.~Ibarz, B.~Ichter, A.~Irpan, E.~Jang, R.~J. Ruano, K.~Jeffrey, S.~Jesmonth, N.~J. Joshi, R.~Julian, D.~Kalashnikov, Y.~Kuang, K.-H. Lee, S.~Levine, Y.~Lu, L.~Luu, C.~Parada, P.~Pastor, J.~Quiambao, K.~Rao, J.~Rettinghouse, D.~Reyes, P.~Sermanet, N.~Sievers, C.~Tan, A.~Toshev, V.~Vanhoucke, F.~Xia, T.~Xiao, P.~Xu, S.~Xu, M.~Yan, and A.~Zeng, ``Do as i can, not as i say: Grounding language in robotic affordances,'' 2022.

\bibitem{NEURIPS2020_1457c0d6}
\BIBentryALTinterwordspacing
T.~Brown, B.~Mann, N.~Ryder, M.~Subbiah, J.~D. Kaplan, P.~Dhariwal, A.~Neelakantan, P.~Shyam, G.~Sastry, A.~Askell, S.~Agarwal, A.~Herbert-Voss, G.~Krueger, T.~Henighan, R.~Child, A.~Ramesh, D.~Ziegler, J.~Wu, C.~Winter, C.~Hesse, M.~Chen, E.~Sigler, M.~Litwin, S.~Gray, B.~Chess, J.~Clark, C.~Berner, S.~McCandlish, A.~Radford, I.~Sutskever, and D.~Amodei, ``Language models are few-shot learners,'' in \emph{Advances in Neural Information Processing Systems}, H.~Larochelle, M.~Ranzato, R.~Hadsell, M.~Balcan, and H.~Lin, Eds., vol.~33.\hskip 1em plus 0.5em minus 0.4em\relax Curran Associates, Inc., 2020, pp. 1877--1901. [Online]. Available: \url{https://proceedings.neurips.cc/paper_files/paper/2020/file/1457c0d6bfcb4967418bfb8ac142f64a-Paper.pdf}
\BIBentrySTDinterwordspacing

\bibitem{wei2023chainofthought}
J.~Wei, X.~Wang, D.~Schuurmans, M.~Bosma, B.~Ichter, F.~Xia, E.~Chi, Q.~Le, and D.~Zhou, ``Chain-of-thought prompting elicits reasoning in large language models,'' 2023.

\bibitem{LeCun2015}
\BIBentryALTinterwordspacing
Y.~LeCun, Y.~Bengio, and G.~Hinton, ``Deep learning,'' \emph{Nature}, vol. 521, no. 7553, pp. 436--444, 2015. [Online]. Available: \url{https://doi.org/10.1038/nature14539}
\BIBentrySTDinterwordspacing

\bibitem{8793485}
M.~A. Lee, Y.~Zhu, K.~Srinivasan, P.~Shah, S.~Savarese, L.~Fei-Fei, A.~Garg, and J.~Bohg, ``Making sense of vision and touch: Self-supervised learning of multimodal representations for contact-rich tasks,'' in \emph{2019 International Conference on Robotics and Automation (ICRA)}, 2019, pp. 8943--8950.

\bibitem{8461249}
T.~Zhang, Z.~McCarthy, O.~Jow, D.~Lee, X.~Chen, K.~Goldberg, and P.~Abbeel, ``Deep imitation learning for complex manipulation tasks from virtual reality teleoperation,'' in \emph{2018 IEEE International Conference on Robotics and Automation (ICRA)}, 2018, pp. 5628--5635.

\bibitem{Zaadnoordijk2022}
\BIBentryALTinterwordspacing
L.~Zaadnoordijk, T.~R. Besold, and R.~Cusack, ``Lessons from infant learning for unsupervised machine learning,'' \emph{Nature Machine Intelligence}, vol.~4, no.~6, pp. 510--520, Jun 2022. [Online]. Available: \url{https://doi.org/10.1038/s42256-022-00488-2}
\BIBentrySTDinterwordspacing

\bibitem{Saxe2021}
\BIBentryALTinterwordspacing
A.~Saxe, S.~Nelli, and C.~Summerfield, ``If deep learning is the answer, what is the question?'' \emph{Nature Reviews Neuroscience}, vol.~22, no.~1, pp. 55--67, Jan 2021. [Online]. Available: \url{https://doi.org/10.1038/s41583-020-00395-8}
\BIBentrySTDinterwordspacing

\bibitem{schulman2017proximal}
J.~Schulman, F.~Wolski, P.~Dhariwal, A.~Radford, and O.~Klimov, ``Proximal policy optimization algorithms,'' 2017.

\bibitem{pmlr-v80-haarnoja18b}
\BIBentryALTinterwordspacing
T.~Haarnoja, A.~Zhou, P.~Abbeel, and S.~Levine, ``Soft actor-critic: Off-policy maximum entropy deep reinforcement learning with a stochastic actor,'' in \emph{Proceedings of the 35th International Conference on Machine Learning}, ser. Proceedings of Machine Learning Research, J.~Dy and A.~Krause, Eds., vol.~80.\hskip 1em plus 0.5em minus 0.4em\relax PMLR, 10--15 Jul 2018, pp. 1861--1870. [Online]. Available: \url{https://proceedings.mlr.press/v80/haarnoja18b.html}
\BIBentrySTDinterwordspacing

\bibitem{10.5555/3294996.3295141}
S.~Gu, T.~Lillicrap, Z.~Ghahramani, R.~E. Turner, B.~Sch\"{o}lkopf, and S.~Levine, ``Interpolated policy gradient: Merging on-policy and off-policy gradient estimation for deep reinforcement learning,'' in \emph{Proceedings of the 31st International Conference on Neural Information Processing Systems}, ser. NIPS'17.\hskip 1em plus 0.5em minus 0.4em\relax Red Hook, NY, USA: Curran Associates Inc., 2017, p. 3849–3858.

\bibitem{10.1145/3170427.3186500}
\BIBentryALTinterwordspacing
N.~Koganti, A.~Rahman H. A.~G., Y.~Iwasawa, K.~Nakayama, and Y.~Matsuo, ``Virtual reality as a user-friendly interface for learning from demonstrations,'' in \emph{Extended Abstracts of the 2018 CHI Conference on Human Factors in Computing Systems}, ser. CHI EA '18.\hskip 1em plus 0.5em minus 0.4em\relax New York, NY, USA: Association for Computing Machinery, 2018, p. 1–4. [Online]. Available: \url{https://doi.org/10.1145/3170427.3186500}
\BIBentrySTDinterwordspacing

\bibitem{10.5555/3454287.3455660}
\BIBentryALTinterwordspacing
Y.~Ding, C.~Florensa, P.~Abbeel, and M.~Phielipp, ``Goal-conditioned imitation learning,'' in \emph{Advances in Neural Information Processing Systems}, H.~Wallach, H.~Larochelle, A.~Beygelzimer, F.~d\textquotesingle Alch\'{e}-Buc, E.~Fox, and R.~Garnett, Eds., vol.~32.\hskip 1em plus 0.5em minus 0.4em\relax Curran Associates, Inc., 2019. [Online]. Available: \url{https://proceedings.neurips.cc/paper/2019/file/c8d3a760ebab631565f8509d84b3b3f1-Paper.pdf}
\BIBentrySTDinterwordspacing

\bibitem{Thor2022}
\BIBentryALTinterwordspacing
M.~Thor and P.~Manoonpong, ``Versatile modular neural locomotion control with fast learning,'' \emph{Nature Machine Intelligence}, vol.~4, no.~2, pp. 169--179, Feb 2022. [Online]. Available: \url{https://doi.org/10.1038/s42256-022-00444-0}
\BIBentrySTDinterwordspacing

\bibitem{doi:10.1126/scirobotics.abb2174}
\BIBentryALTinterwordspacing
C.~Yang, K.~Yuan, Q.~Zhu, W.~Yu, and Z.~Li, ``Multi-expert learning of adaptive legged locomotion,'' \emph{Science Robotics}, vol.~5, no.~49, p. eabb2174, 2020. [Online]. Available: \url{https://www.science.org/doi/abs/10.1126/scirobotics.abb2174}
\BIBentrySTDinterwordspacing

\bibitem{bellemare2016unifying}
M.~G. Bellemare, S.~Srinivasan, G.~Ostrovski, T.~Schaul, D.~Saxton, and R.~Munos, ``Unifying count-based exploration and intrinsic motivation,'' 2016.

\bibitem{ostrovski2017count}
G.~Ostrovski, M.~G. Bellemare, A.~van~den Oord, and R.~Munos, ``Count-based exploration with neural density models,'' in \emph{Proceedings of the 34th International Conference on Machine Learning - Volume 70}, ser. ICML'17.\hskip 1em plus 0.5em minus 0.4em\relax JMLR.org, 2017, p. 2721–2730.

\bibitem{9492850}
E.~Triantafyllidis, W.~Hu, C.~McGreavy, and Z.~Li, ``Metrics for 3d object pointing and manipulation in virtual reality: The introduction and validation of a novel approach in measuring human performance,'' \emph{IEEE Robotics Automation Magazine}, pp. 2--17, 2021.

\bibitem{juliani2018unity}
A.~Juliani, V.-P. Berges, E.~Teng, A.~Cohen, J.~Harper, C.~Elion, C.~Goy, Y.~Gao, H.~Henry, M.~Mattar, \emph{et~al.}, ``Unity: A general platform for intelligent agents,'' \emph{arXiv preprint arXiv:1809.02627}, 2018.

\bibitem{8202133}
J.~Tobin, R.~Fong, A.~Ray, J.~Schneider, W.~Zaremba, and P.~Abbeel, ``Domain randomization for transferring deep neural networks from simulation to the real world,'' in \emph{2017 IEEE/RSJ International Conference on Intelligent Robots and Systems (IROS)}, 2017, pp. 23--30.

\bibitem{Billardeaat8414}
\BIBentryALTinterwordspacing
A.~Billard and D.~Kragic, ``Trends and challenges in robot manipulation,'' \emph{Science}, vol. 364, no. 6446, 2019. [Online]. Available: \url{https://science.sciencemag.org/content/364/6446/eaat8414}
\BIBentrySTDinterwordspacing

\bibitem{10.5555/3304652.3304697}
F.~Torabi, G.~Warnell, and P.~Stone, ``Behavioral cloning from observation,'' in \emph{Proceedings of the 27th International Joint Conference on Artificial Intelligence}, ser. IJCAI'18.\hskip 1em plus 0.5em minus 0.4em\relax AAAI Press, 2018, p. 4950–4957.

\bibitem{NIPS2016_cc7e2b87}
\BIBentryALTinterwordspacing
J.~Ho and S.~Ermon, ``Generative adversarial imitation learning,'' in \emph{Advances in Neural Information Processing Systems}, D.~Lee, M.~Sugiyama, U.~Luxburg, I.~Guyon, and R.~Garnett, Eds., vol.~29.\hskip 1em plus 0.5em minus 0.4em\relax Curran Associates, Inc., 2016. [Online]. Available: \url{https://proceedings.neurips.cc/paper/2016/file/cc7e2b878868cbae992d1fb743995d8f-Paper.pdf}
\BIBentrySTDinterwordspacing

\bibitem{Wolpert2011}
\BIBentryALTinterwordspacing
D.~M. Wolpert, J.~Diedrichsen, and J.~R. Flanagan, ``Principles of sensorimotor learning,'' \emph{Nature Reviews Neuroscience}, vol.~12, no.~12, pp. 739--751, Dec 2011. [Online]. Available: \url{https://doi.org/10.1038/nrn3112}
\BIBentrySTDinterwordspacing

\bibitem{Osband2020Behaviour}
\BIBentryALTinterwordspacing
I.~Osband, Y.~Doron, M.~Hessel, J.~Aslanides, E.~Sezener, A.~Saraiva, K.~McKinney, T.~Lattimore, C.~Szepesvari, S.~Singh, B.~V. Roy, R.~Sutton, D.~Silver, and H.~V. Hasselt, ``Behaviour suite for reinforcement learning,'' in \emph{International Conference on Learning Representations}, 2020. [Online]. Available: \url{https://openreview.net/forum?id=rygf-kSYwH}
\BIBentrySTDinterwordspacing

\bibitem{Triantafyllidis2024Intrinsic}
E.~Triantafyllidis, F.~Christianos, and Z.~Li, ``Intrinsic language-guided exploration for complex long-horizon robotic manipulation tasks,'' \url{https://www.youtube.com/watch?v=Z0NHHs2IfPw}, 2024, video presentation accepted at the International Conference on Robotics and Automation (ICRA) 2024, Yokohama, Japan.

\bibitem{radford2021learning}
A.~Radford, J.~W. Kim, C.~Hallacy, A.~Ramesh, G.~Goh, S.~Agarwal, G.~Sastry, A.~Askell, P.~Mishkin, J.~Clark, G.~Krueger, and I.~Sutskever, ``Learning transferable visual models from natural language supervision,'' 2021.

\end{thebibliography}

\end{document}